\documentclass[lettersize,journal]{IEEEtran}
\usepackage{amsmath,amsfonts}
\usepackage{array}
\usepackage[caption=false,font=normalsize,labelfont=sf,textfont=sf]{subfig}
\usepackage{textcomp}
\usepackage{stfloats}
\usepackage{url}
\usepackage{verbatim}
\usepackage{graphicx}
\def\BibTeX{{\rm B\kern-.05em{\sc i\kern-.025em b}\kern-.08em
    T\kern-.1667em\lower.7ex\hbox{E}\kern-.125emX}}
\usepackage{balance}

\usepackage{graphicx}
\usepackage{multirow}
\usepackage{amsmath,amssymb,amsfonts}
\usepackage{amsthm}
\usepackage{mathrsfs}
\usepackage{xcolor}
\usepackage{textcomp}
\usepackage{manyfoot}
\usepackage{booktabs}
\usepackage{algorithm}
\usepackage{algorithmicx}
\usepackage{algpseudocode}
\usepackage{listings}
\usepackage{makecell}
\usepackage{url}

\begin{document}
\title{Clustering-Guided Multi-Layer Contrastive Representation Learning for Citrus Disease Classification}
\author{Jun Chen, Yonghua Yu, Weifu Li, Yaohui Chen, Hong Chen
\thanks{J. Chen, W. Li, and H. Chen are with the College of Informatics, Huazhong Agricultural University, Wuhan 430070, China (email: cj850487243@163.com, liweifu@mail.hzau.edu.cn, chenh@mail.hzau.edu.cn).

W. Li and H. Chen are also with the Engineering Research Center of Intelligent Technology for Agriculture, Ministry of Education, Wuhan, 430070, Hubei, China.

Y. Yu and Y. Chen are with College of Engineering, Huazhong Agricultural University, Wuhan, 430070, Hubei, China (email: yonghuayu@webmail.hzau.edu.cn , yaohui.chen@mail.hzau.edu.cn)

Y. Chen is also with the National Key Laboratory for Germplasm Innovation \& Utilization of Horticultural Crops, Wuhan, 430070, Hubei, China.

H. Chen is the corresponding author.}}

\markboth{Journal of \LaTeX\ Class Files,~Vol.~18, No.~9, September~2020}%
{How to Use the IEEEtran \LaTeX \ Templates}

\maketitle

\begin{abstract}
Citrus, as one of the most economically important fruit crops globally, suffers severe yield depressions due to various diseases. 
Accurate disease detection and classification serve as critical prerequisites for implementing targeted control measures. 
Recent advancements in artificial intelligence, particularly deep learning-based computer vision algorithms, have substantially decreased time and labor requirements while maintaining the accuracy of detection and classification. 
Nevertheless, these methods predominantly rely on massive, high-quality annotated training examples to attain promising performance. 
By introducing two key designs: contrasting with cluster centroids and a multi-layer contrastive training (MCT) paradigm, this paper proposes a novel clustering-guided self-supervised multi-layer contrastive representation learning (CMCRL) algorithm. 
The proposed method demonstrates several advantages over existing counterparts: (1) optimizing with massive unannotated samples; (2) effective adaptation to the symptom similarity across distinct citrus diseases; (3) hierarchical feature representation learning. 
The proposed method achieves state-of-the-art performance on the public citrus image set CDD, outperforming existing methods by 4.5\%-30.1\% accuracy. 
Remarkably, our method narrows the performance gap with fully supervised counterparts (all samples are labeled). 
Specifically, CMCRL achieves 93.75\% accuracy on CDD and 92.79\% on our proposed citrus disease image dataset CitrusDisease7, compared to 92.19\% and 93.27\%, respectively, from fully supervised learning. 
The ablation experiments under various encoder models and data augmentation strategies validate the generalizability of the proposed CMCRL. 
Beyond classification accuracy, our method shows great performance on other evaluation metrics (F1 score, precision, and recall), highlighting the robustness against the class imbalance challenge.
\end{abstract}

\begin{IEEEkeywords}
Citrus disease classification, symptom similarity, self-supervised contrastive learning, density clustering, deep learning.
\end{IEEEkeywords}

\section{Introduction}
\label{introduction}
Citrus fruit, one of the world's most extensively cultivated fruit crops, is believed that it was spread from Southeast Asia to other regions of the world, including East Asia, North America, South America, and Europe \cite{MatheyambathPP2016,CasertaGD2019}. 
It enjoys rich and low-calorie nutrients (mainly vitamin C, fiber, potassium, and folate) \cite{JiaZO2017}.
Currently, multifaceted constraints brought by biotic stresses (pathogens and pests) and abiotic stresses (temperature, drought, soil salinity) collectively limit the production of citrus fruit, triggering significant socioeconomic repercussions \cite{Syed-Ab-RahmanHP2022,SunNK2019}. 
For example, the estimated loss caused by a devastating citrus disease, huanglongbing (HLB), between 2016 and 2020 surpasses one billion per year in Florida, with the elimination of approximately five thousand jobs annually \cite{LiWD2020}. 
\cite{LopezD2014} estimated the potential economic loss due to HLB would be $\$$2.7 billion USD without any management over 20 years in California. 
Except for huanglongbing, citrus production is threatened by multiple diseases including canker \cite{Das2003}, anthracnose \cite{TangYZ2023}, black spot \cite{Martínez-MinayaCL2015}, melanose \cite{ZhouLL2024}, and sunscald \cite{ParkKY2022}.

These citrus diseases are mainly caused by three kinds of factors, i.e., bacterium (huanglongbing \cite{Syed-Ab-RahmanHP2022,BaoLC2023} and canker \cite{MartinsdB2020}), fungi (anthracnose \cite{AielloCG2015}, black spot \cite{Benson1895}, and melanose \cite{ChaisiriLL2020}), and abiotic factors (sunscald \cite{ParkKY2022}). 
Many effective measures were proposed to control the transmission of diseases. 
For instance, insecticides, copper, and prompt removal of infected plant issues are utilized to inhibit bacterial reproduction \cite{GrahamGC2004,BoinaBC2015,ZhangLJ2025}.
Besides, the approaches are mainly based on physical, chemical, biological, and biotechnology treatments such as vapor heat, synthetic fungicides (mancozeb and prochloraz), \textit{Bacillus subtilis}, and genome editing, respectively \cite{YiHH2014,CiofiniNB2022,Syed-Ab-RahmanHP2022}.
As for abiotic diseases, some necessary abiotic factors, such as temperature, humidity, sunlight, and soil, are required to be controlled within their corresponding appropriate ranges \cite{ParkKY2022}. 

Despite the existence of various interventions, accurate disease detection and classification constitute the prerequisite for cost-effective disease management. 
\cite{MartinelliSD2015} listed two traditional molecular methods for disease detection: serological assays (enzyme-linked immunosorbent assay) and nucleic acid-based methods (polymerase chain reaction, PCR).
\cite{TianZC2020} further stated that the PCR assay is performed for early plant disease diagnosis even when symptoms are not visible. 
However, these conventional techniques face inherent constraints: the complicated sampling procedures and the limited plant applicability \cite{TangYZ2023}. 
Recent advances in machine learning have driven significant methodological innovations for rapid and accurate plant disease diagnosis, detection, and classification \cite{SaleemPA2019,SinghSS2020,LoeyEA2020,AhmadSG2023}. 
Citrus disease research has particularly benefited from machine learning, even deep learning applications \cite{DengLH2016,SharifKI2018,AbdulridhaBA2019,TangYZ2023}.
Current supervised learning paradigms achieve optimal accuracy when trained on extensive annotated datasets \cite{KimKB2022}. 
The laborious and expensive process of collecting labeled plant disease data, as well as the limited amount of data in most public datasets \cite{NazkiYF2020}, underscores the critical importance of studying how to fully utilize unlabeled data \cite{ZhaoZL2023,ChaiLT2024,YilmaDA2025}. 
Although a few works have made some related attempts, the symptom similarity among these citrus diseases poses challenges to accurate detection and classification \cite{FangLL2021,MonowarHK2022}. 

This paper is devoted to exploring the utilization of unlabeled citrus images for citrus disease classification task. 
The main contributions are summarized as follows:
\begin{itemize}
    \item Considering that some symptoms of citrus diseases are challenging to be distinguished without expert knowledge, we introduce the self-supervised contrastive pre-training combined with cluster pre-preprocessing (Cluster Contrast \footnote{Cluster Contrast was originally proposed for fine-grained visual classification task, person re-identification \cite{DaiWY2022}. Our method replaces the linear projection of Cluster Contrast with non-linear projection (NP) heads.}) to the citrus disease classification task. 
    This pre-training process utilizes massive unannotated samples to implement contrastive learning with cluster centroids rather than the contrast among samples, which effectively adapts to the symptom similarity across distinct citrus diseases. 

    \item To obtain better embedding feature vectors, we propose a novel \textbf{C}lustering-guided \textbf{M}ulti-layer \textbf{C}ontrastive \textbf{R}epresentation \textbf{L}earning (CMCRL) algorithm via further developing a \textbf{M}ulti-layer \textbf{C}ontrastive \textbf{T}raining (MCT) paradigm. 
    This paradigm exploits the hierarchical feature structural information via the integration of multiple intermediate feature maps. 

    \item Extensive evaluations on the public dataset CDD and our proposed dataset CitrusDisease7 validate the effectiveness and broad applicability of our method. 
    Experimental results show that CMCRL can achieve the classification performance comparable to fully supervised classification \footnote{In the fully supervised case, the unlabeled pre-training samples in the self-supervised case are all labeled.}. 
    Besides, CMCRL can improve the model performance on unbalanced dataset. 
\end{itemize}

The rest of this paper is organized as follows. 
Section \ref{related works} reviews the related works on plant disease detection and classification based on machine learning. 
Section \ref{materials and method} demonstrates the datasets, proposed method, and evaluation metrics. 
Experimental results and analyses are provided in Section \ref{results}. 
Section \ref{conclusions} concludes this work and gives future work.

\section{Related works}
\label{related works}
In this section, we will introduce some works related to plant disease detection and classification based on traditional machine learning and deep learning including self-supervised learning. 

\subsection{Plant disease detection and classification based on machine learning}
For traditional machine learning-related work, \cite{RumpfMS2010} combined Support Vector Machines (SVM) and spectral vegetation indices to early detect and differentiate sugar beet diseases. 
They achieved three aims: 1) the distinguishment between diseased and non-diseased leaves (97$\%$ accuracy); 2) the classification between Cercospora leaf spot, leaf rust, and powdery mildew (86$\%$ accuracy); and 3) the disease identification before visible symptoms appear (accuracy between 65$\%$ and 90$\%$).
\cite{CaoLZ2015} found that the area of the red edge peak is more sensitive for powdery mildew detection than others like difference vegetation index. 
They empirically showed that partial least square regression can more accurately estimate the disease index of powdery mildew and grain yield. 
\cite{AdeelKS019} designed a system comprising four steps for grape leaf disease diagnosis and recognition. 
Specifically, the first step increases the local contrast of symptoms via a local contrast haze reduction. 
Then, the second step selects the best LAB color channel which will be fused by canonical correlation analysis in the third step. 
Finally, the classification of features removing noise is then conducted by M-class SVM. 

As for deep learning-related work, \cite{WangVH2019} proposed a method based on generative adversarial nets to achieve 96.25$\%$ classification accuracy. 
Similarly, for tomato disease detection, \cite{SharmaBG2020} used segmented tomato disease images to train the convolutional neural network (CNN), which achieved 98.6$\%$ accuracy.
Instead of randomly initializing model parameters, \cite{ChenCZ2020} adopted the pre-trained VGGNet learned from a popular large labeled dataset, ImageNet. 
They added an Inception module at the end of VGGNet, which is trained to transfer the performance on ImageNet to their rice disease class prediction task, achieving 92.00$\%$ accuracy. 

\subsection{Plant disease detection and classification based on self-supervised learning}
In recent years, some literature has made some attempts to save the cost caused by manual annotation. 
\cite{FangLL2021} designed a self-supervised clustering framework for plant disease image classification. 
It consists of Kernel K-means producing pseudo labels and a deep learning module further classifying samples. 
The experiments on two public datasets, PlantVillage \cite{HughesS2016} and Citrus Disease Dataset (CDD, \cite{RaufSL2019}), showed high superiority compared with previous classification algorithms. 
\cite{MonowarHK2022} also introduced a self-supervised clustering system to conduct leaf disease identification. 
The first part of this system is an embedding siamese convolutional neural network (CNN) trained with AutoEmbedder. 
Then, the final classification result is obtained via K-means. 
Their empirical performance is better on PlantVillage and CDD compared to \cite{FangLL2021}. 
\cite{KimKB2022} utilized a self-supervised pre-training feature extractor to enhance feature representation. 
\cite{ChaiLT2024} incorporated self-supervised learning into a previous model (FF-ViT, \cite{HaoChaiHT2023}) to introduce a novel model, Cross Learning Vision Transformer (CL-ViT), which enhances conceptual feature disentanglement. 
\cite{WangYL2024} adopted a masked autoencoder to alleviate the requirement of a large amount of labeled data for crop disease recognition. 

\subsection{Citrus disease detection and classification based on machine learning}
For citrus diseases focused on by this work, \cite{DengLH2016} employed the support vector classification method to detect citrus HLB. 
\cite{SharifKI2018} first selected hybrid features via covariance based on primary component analysis score, entropy, and skewness, and then classified various citrus diseases with support vector machine. 
\cite{AbdulridhaBA2019} used two methods, radial basis function and K nearest neighbor, respectively, to detect citrus canker in several disease stages. 
\cite{PartelNS2019} developed a software using two deep CNNs to monitor Asian citrus psyllid which is a key vector of HLB.
\cite{FangLL2021} and \cite{MonowarHK2022} conducted some experiments on citrus data with their self-supervised learning models. 
They indicated that their accuracy is low compared to supervised methods. 

This paper introduces a new clustering-guided self-supervised training paradigm to bridge the performance gap between self-supervised models and supervised models for the citrus disease classification task.

\begin{figure*}[ht]
    \centering
    \includegraphics[width=\textwidth]{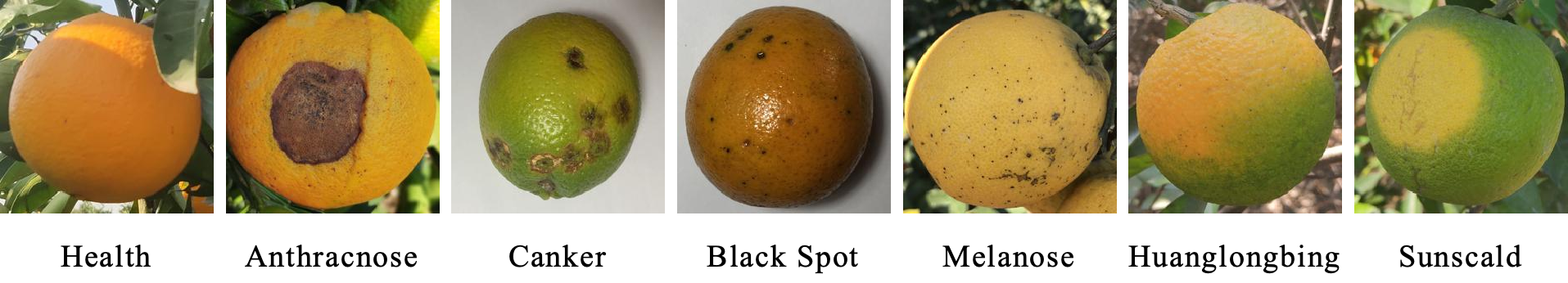}
    \caption{Samples of 7 classes citrus diseases taken from CitrusDisease7.}
    \label{CitrusDisease7}
\end{figure*}

\begin{figure}[!t]
    \centering
    \includegraphics[width=0.47\textwidth]{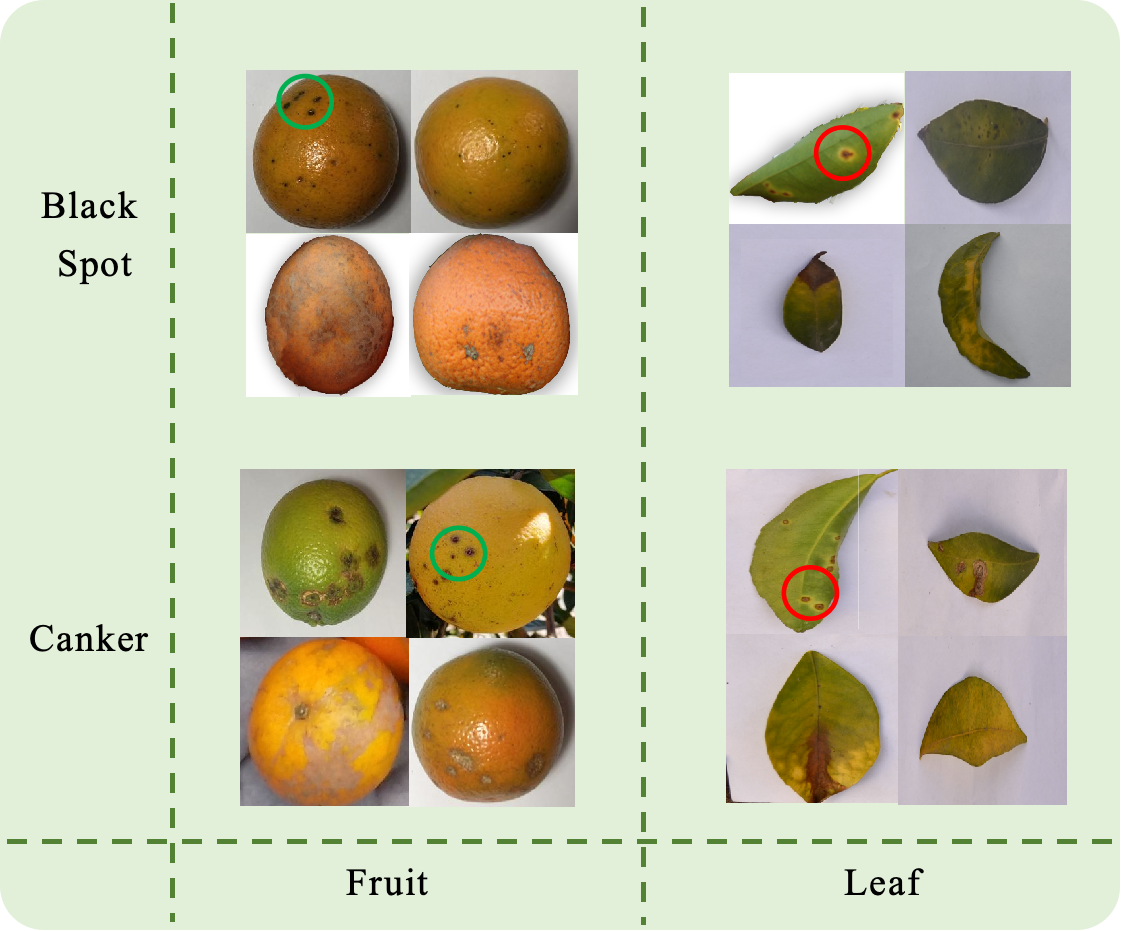}
    \caption{Phenotypic comparisons between citrus black spot and canker on fruit and leaf. For example, green and red circles indicate two kinds of similar phenotypes of fruit and leaf, respectively.}
    \label{Black spot vs canker}
\end{figure}

\section{Materials and method}
\label{materials and method}

\subsection{Datasets and pre-processing}
In this work, we construct a citrus fruit disease image dataset (CitrusDisease7) comprising in-situ collected and web-sourced\footnote{\url{http://www.dilitanxianjia.com/1762/}} citrus pathology images. 
Our dataset contains 3956 images of healthy citrus and 6 kinds of citrus diseases (as shown in Table \ref{Datasets} and Figure \ref{CitrusDisease7}), where 3161 images are regarded as unlabeled pre-training samples\footnote{Although these 3161 images are labeled, the label information is hidden during pretraining.}, 594 fine-tuning images, and 201 testing images. 

In addition, we conduct experiments on a public dataset (CDD \cite{RaufSL2019} shown in Table \ref{Datasets}), making comparisons with existing approaches. 
Following previous works, we use the leaf image subset of CDD, allocating 485 images for pre-training and 124 for testing. 
Note that CDD exhibits a smaller scale and a relatively mild class imbalance ratio compared to CitrusDisease7. 

\begin{table}[!t]
\caption{Details of citrus disease datasets used by this work.}
\label{Datasets}
    \centering
    \renewcommand\arraystretch{1.5}
    \begin{tabular}{@{\hspace{0.3cm}}c@{\hspace{0.3cm}}@{\hspace{0.3cm}}c@{\hspace{0.3cm}}@{\hspace{0.3cm}}c@{\hspace{0.3cm}}@{\hspace{0.3cm}}c@{\hspace{0.3cm}}}
        \hline
        Dataset & Organ & Disease/Health & Amount\\
        \hline
        \multirow{7}{*}{CitrusDisease7} & \multirow{7}{*}{Fruit} & anthracnose & 58\\
        & & blackspot & 1236\\
        & & canker & 1707\\
        & & huanglong & 86\\
        & & melanose & 327\\
        & & health & 442\\
        & & sunscald & 100\\
        \hline
        \multirow{5}{*}{CDD} & \multirow{5}{*}{Leaf} & blackspot & 171\\
        & & canker & 163\\
        & & huanglong & 204\\
        & & health & 58\\
        & & melanose & 13\\
        \hline
    \end{tabular}
\end{table}

\subsection{Overview of proposed method}
As far as we know, current self-supervised approaches for citrus disease classification \cite{FangLL2021, MonowarHK2022} exhibit two critical limitations: (1) suboptimal accuracy relative to supervised counterparts; (2) high dependence on the consistency between latent labels and clustering pseudo labels. 
Since the symptom similarity of citrus diseases shown in Figure \ref{Black spot vs canker}, it is challenging to achieve highly accurate clustering. 
To address these limitations, this paper introduce a new self-supervised contrastive learning paradigm, CMCRL. 
Our code is provided in https://github.com/cjml0808/CMCRL.

\begin{figure*}[ht]
    \centering
    \includegraphics[width=\textwidth]{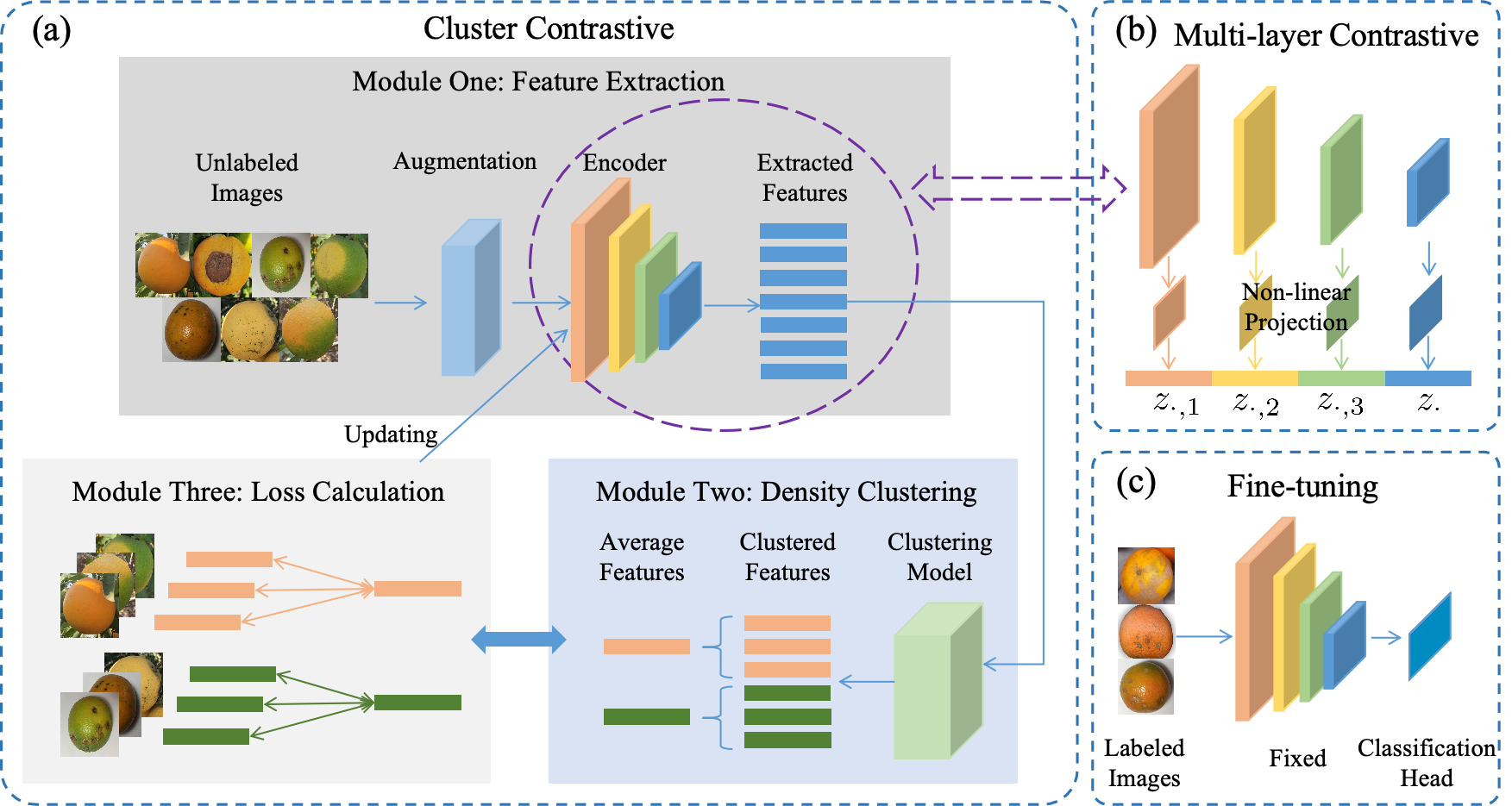}
    \caption{Overview of the proposed CMCRL. (a) The framework of Cluster Contrastive; (b) the new multi-layer contrastive training paradigm, MCT; (c) the supervised fine-tuning process with fixed encoder and the linear classification head.}
    \label{Figure of pipeline}
\end{figure*}

As presented in Figure \ref{Figure of pipeline} (a), the pipeline of Cluster Contrast consists of three modules: feature extraction (Module One), density clustering (Module Two), and loss calculation (Module Three). 
In Module One, all unlabeled images are first transformed via some augmentation strategies before being encoded into latent representations via a backbone encoder. 
Module Two clusters all latent representations using a classic density clustering algorithm, DBSCAN. 
A memory dictionary is constructed to store all cluster centroids. 
Module Three calculates contrastive loss between instance-level clustered vectors and their corresponding cluster centroids. 
To enhance feature learning, we design a new multi-layer contrastive training paradigm (MCT, Figure \ref{Figure of pipeline} (b)) optimized with a multi-layer contrastive loss (ML-InfoNCE). 
Except for the final output embedding vector, several intermediate feature maps are also extracted as contrastive representations. 
Finally, the pre-trained encoder remains frozen, while the linear classification head is fine-tuned on a small set of labeled images (Figure \ref{Figure of pipeline} (c)). 
In summary, the pipeline of our new algorithm (CMCRL) is composed of the new module, Module Two, Module Three, and fine-tuning. 

\begin{figure*}[ht]
    \centering
    \includegraphics[width=\textwidth]{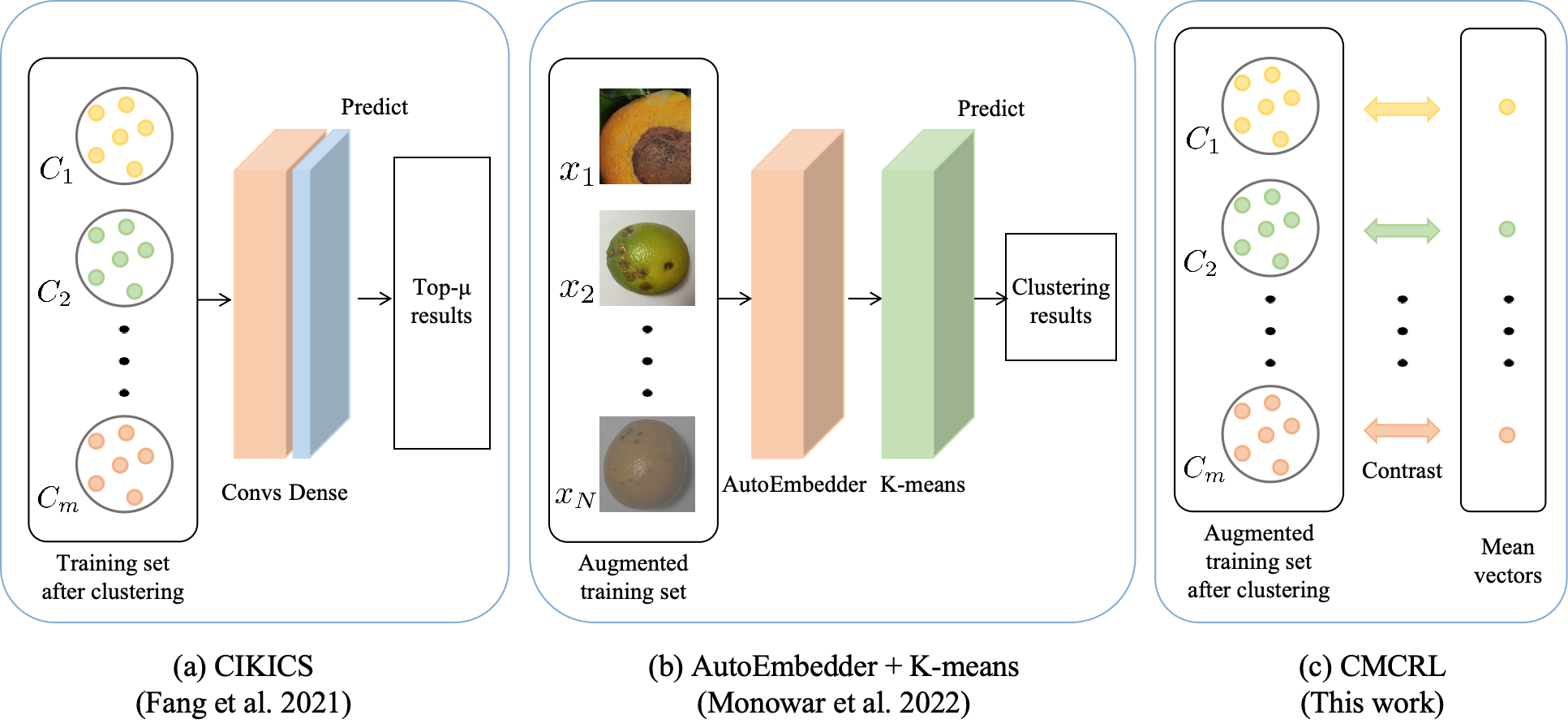}
    \caption{Comparison of three self-supervised training processes based on clustering. $C_j, j\in [m]$ denotes the $j$-th cluster group where the pseudo labels $\tilde{y}$ of all vectors are the same.}
    \label{CMCRL vs previous works}
\end{figure*}

\noindent \textbf{(1) Module One}

This module is employed to extract the embedding vectors for the samples transformed via image augmentation strategy. 
Specifically, we mainly implement a variant of Resnet, Resnet50-IBN, which integrates instance normalization (IN) and batch normalization (BN). 
As \cite{PanLS2018} disclosed, IN can ensure visual invariance, and BN can preserve content-related features and accelerate training. 
Thus, Resnet50-IBN is more suitable for the challenge indicated by Figure \ref{Black spot vs canker} than some traditional backbone networks such as Resnet18, Resnet50, and Resnet152 \cite{HeZR2016}, which is validated by Tables \ref{Cluster Contrast vs NP vs MCT with ResNet50-IBN on CitrusDisease7} and \ref{Cluster Contrast vs NP vs MCT with other encoders on CitrusDisease7}. 

In this paper, we denote the unlabeled pre-training image set by $X = \{x_1, ..., x_N\}, x_i\in \mathbb{R}^{W\times H \times C}, i\in [N] = \{1,...,N\}$, where $N, W, H$, and $C$ are the count of images and the width, height, and the number of channels for image. 
We assume the latent ground-truth labels are $\{y_i\}_{i=1}^N, y_i\in [K]$. 
$X$ is generated through the augmentation $\mathcal{T}$ which comprises random crop (RC), random erase (RE), padding (Pad), and random horizontal flipping (RHF). 
Subsequently, the encoder $f: \mathbb{R}^{W\times H \times C} \rightarrow \mathbb{R}^d$ extracts the corresponding feature vector set $Z = \{z_i\}_{i=1}^N, z_i\in \mathbb{R}^d$, where the embedding dimension $d$ is manually set. 

\noindent \textbf{(2) Module Two}

This module first applies DBSCAN to produce pseudo labels, denoted as $\{\tilde{y}_i\}_{i=1}^N$, for embedded representations. 
Note that unclustered samples are assigned the exclusion marker $-1$, which will be discarded in Module Three. 
Let $m$ be the number of cluster groups \footnote{The cluster count $m$ remains constant in each training epoch and is dynamically updated upon DBSCAN re-clustering in the subsequent epoch.}, then the pseudo labels satisfy $\tilde{y}_i\in \{-1, 1, ..., m\}$. 
Finally, the cluster centroid $\bar{z}_j$ ($j\in [m]$) for each cluster group $C_j\in \{C_1, ..., C_m\}$ is computed using these pseudo labels as 
\begin{align}
\label{cluster centroid}
    \bar{z}_j = \sum_{z_i\in C_j} z_i / |C_j|,
\end{align}
where $|C_j|$ is the number of samples assigned to cluster $C_j$, let $\sum_{j=1}^m |C_j| = N_C$. 
These centroids are subsequently stored in a memory dictionary. 
In each epoch, these cluster centroids are updated with the momentum $\alpha$ as follows: 
\begin{align}
\label{momentum update}
    \bar{z}_j = \alpha \bar{z}_j + (1-\alpha) z_{C_j},
\end{align}
where $z_{C_j}$ is the sample feature belonging to $C_j$ in a data batch. 

\begin{algorithm*}
\caption{Pseudo Code of CMCRL}
\label{Pseudo code of CMCRL}
\begin{algorithmic}[1]
\Require Unlabeled augmented image set $X$, density clustering algorithm DBSCAN, initial encoder $f_{0,0}$, temperature parameter $\tau$, momentum parameter $\alpha$, batch size $B$, number of epoch $T$, number of iteration $S$
\Ensure Pre-trained encoder $f_{T,S}$
\For{t=1,...,T}
    \State Obtain the feature sets $\{Z_1, Z_2, Z_3, Z\} = f_{t-1,S}(X)$ from multiple layers and let $f_{t,0}=f_{t-1,S}$
    \State Use DBSCAN to cluster $N$ samples into $m$ clusters $\{C_j\}_{j=1}^m$ based on $\{Z_1, Z_2, Z_3, Z\}$, satisfying $\sum_{j=1}^m |C_j| = N_C \leq N$ and $C_j\cap C_{j^\prime} = \varnothing$ for $\forall j, j^\prime\in[m]$
    \State Compute the cluster centroids $\{\bar{z}_j\}_{j=1}^m$ with Equation \eqref{cluster centroid} and regard them as the initialization of memory dictionary 
    \For{s=1,...,S}
        \State Randomly select $B$ samples to compute MLNCE loss with Equation \eqref{MLNCE}
        \State Update the encoder $f_{t,s-1}$ to $f_{t,s}$ by optimizer
        \State Update memory dictionary with Equation \eqref{momentum update}
    \EndFor
\EndFor
\end{algorithmic}
\end{algorithm*}

\noindent \textbf{(3) Module Three}

Existing clustering-based self-supervised frameworks \cite{FangLL2021,MonowarHK2022} directly leverage cluster assignments as supervisory signals or final predictions. 
The advantage of Module Three lies in contrasting the clustered representations against their cluster centroids, making it less dependent on clustering results. 
The comparison of the self-supervised training processes for CMCRL and previous methods is shown in Figure \ref{CMCRL vs previous works}.

\noindent \textbf{(4) CMCRL}

\cite{ZeilerF2014} had some findings: (1) shallow layers mainly learn basic visual features such as edges, colors; (2) middle layers capture more complex structures such as geometric shapes; (3) deep layers learn high-level semantic features such as entire objects. 
Therefore, the proposed CMCRL algorithm extends the Cluster Contrast framework by integrating the MCT component. 
This development incorporates the feature maps of multiple intermediate layers to exploit hierarchical feature information. 
Specifically, we compute the multi-layer contrastive loss, MLNCE, using the combination of multiple feature maps from different layers. 
It is defined as Equation \eqref{MLNCE}, developing from the InfoNCE loss \cite{HeFW2020}: 
\begin{align}
\label{MLNCE}
    & L_{MLNCE}\notag\\
    = & -\log \left(\frac{\exp\left(\sum_{k=1}^4 z_{\cdot,k}^\top\cdot \bar{z}_{\cdot,k}^+/\tau\right)}{\sum_{j=1}^m\exp\left(\sum_{k=1}^4 z_{\cdot,k}^\top\cdot \bar{z}_{j,k}/\tau\right)} \right),
\end{align}
where $z_{\cdot,k} (k\in [4])$ denotes the feature map of the $k$-th layer, $z_{\cdot,4}=z_{\cdot}\in \{z_1,...,z_N\}$, $\bar{z}_{\cdot,k}^+$ is the cluster centroid with the pseudo label of $z_{\cdot}$, $\bar{z}_{j,k}$ is the cluster centroid of the $j$-th cluster for the $k$-th layer, and $\tau$ represents a temperature parameter. 
The self-supervised pre-training process is detailed presented in Algorithm \ref{Pseudo code of CMCRL}. 

\noindent \textbf{(5) Fine-tuning Process}

The self-supervised pretraining phase is succeeded by supervised fine-tuning using small-scale labeled samples for the adaptation of downstream task. 
We denote the small-scale target image set by $X^\prime = \{x_1^\prime, ..., x_n^\prime\}$ with the labels $\{y_i^\prime\}_{i=1}^n$, where $x_i^\prime\in \mathbb{R}^{W\times H\times C}, y_i^\prime\in [K], i\in [n]$. 
In this work, we adopt the strategy that pre-training encoder remains frozen to preserve learned representations, while fine-tuning the linear classification head $g: \mathbb{R}^d \rightarrow \mathbb{R}^K$.


\begin{table*}[!t]
\caption{Comparison of clustering performance and downstream citrus disease classification accuracy for different methods on the dataset CDD. The best results are marked in bold.}
\label{CMCRL vs CIKICS vs AutoEmbedder on CDD}
    \centering
    \setlength{\tabcolsep}{1mm}
    \renewcommand\arraystretch{1.5}
    \begin{tabular}{@{\hspace{0.5cm}}c@{\hspace{0.5cm}}|@{\hspace{0.5cm}}c@{\hspace{0.5cm}}|@{\hspace{1cm}}c@{\hspace{1cm}}c@{\hspace{1cm}}c@{\hspace{1cm}}}
        \hline
        Method & Clustering or not & CACC & ARI & ACC\\
        \hline
        BYOL \cite{GrillSA2020} & No & $-$ & $-$ & 0.637\\
        SimSiam \cite{ChenH2021} & No & $-$ & $-$ & 0.742\\
        SimCLR \cite{ChenKN2020} & No & $-$ & $-$ & 0.750\\
        CIKICS \cite{FangLL2021} & Yes & 0.582 & 0.369 & 0.641\\
        AutoEmbedder + K-means \cite{MonowarHK2022} & Yes & \textbf{0.851} & \textbf{0.766} & 0.893\\
        Cluster Contrast & Yes & 0.540 & 0.150 & 0.891\\
        Cluster Contrast + NP & Yes & 0.652 & 0.238 & 0.922\\
        Cluster Contrast + MCT (CMCRL) & Yes & 0.600 & 0.228 & \textbf{0.938}\\
        \hline
    \end{tabular}
\end{table*}

\subsection{Evaluation metrics}

Considering the combination of contrastive learning and clustering algorithm, this work adopts two sets of evaluation metrics. 

The first metric set is composed of several clustering metrics including cluster accuracy (CACC) and adjusted random index (ARI).
CACC is defined as 
\begin{align*}
    \mathrm{CACC} = \max_{\sigma} \frac{\sum_{q=1}^{N_C} \mathbb{I} \left( y_{i_q}=\sigma(\tilde{y}_{i_q})\right)}{N_C}, 
\end{align*}
where $\sigma$ ranges over all possible one-to-one mapping of the labels and clusters \cite{OhiMS2020}, $\mathbb{I}$ is the indicator function, $\{i_1,...,i_q\}$ is an arbitrary permutation of the indexes for all clustering samples. 
ARI is defined as 
\begin{align*}
    \mathrm{ARI} = \frac{\sum_{ij} \binom{N_{ij}}{2} - \frac{\sum_i \binom{a_i}{2} \sum_j \binom{b_j}{2}}{\binom{N_{ij}}{2}}}{\frac{\sum_i \binom{a_i}{2} + \sum_j \binom{b_j}{2}}{2} - \frac{\sum_i \binom{a_i}{2} \sum_j \binom{b_j}{2}}{\binom{N_{ij}}{2}}},
\end{align*}
where $N_{ij}$ is the number of the samples with the ground-truth label $y=i$ and the pesudo label $\tilde{y}=j$, $a_i = \sum_{j=1}^m N_{ij}, b_j = \sum_{i=1}^{K} N_{ij}, \binom{N_{ij}}{2} = \frac{N_{ij} (N_{ij}-1)}{2}, i,j\in [K]$. 

The second metric set includes four classification metrics, i.e., top-1 accuracy (ACC), Recall, Precision, and F1 score. 
We define them as 
\begin{align*}
    \frac{1}{n} \sum_{i=1}^n \mathbb{I}\left(y_i^\prime = \arg\max_{j\in[K]} g\left(f\left(x_i^\prime\right)\right)_j\right),
\end{align*}
\begin{align*}
    \frac{1}{K} \sum_{q\in[K]} \frac{\#\{x_i^\prime| y_i^\prime=q, \arg\max\limits_{j\in[K]} g(f(x_i^\prime))_j=q\}}{\#\{x_i^\prime| y_i=q\}},
\end{align*}
\begin{align*}
    \frac{1}{K} \sum_{q\in[K]} \frac{\#\{x_i^\prime| y_i^\prime=q, \arg\max\limits_{j\in[K]} g(f(x_i^\prime))_j=q\}}{\#\{x_i^\prime| \arg\max\limits_{j\in[K]} g(f(x_i^\prime))_j=q\}},
\end{align*}
and 
\begin{align*}
    \frac{2\times \mathrm{Precision} \times \mathrm{Recall}}{\mathrm{Precision + Recall}},
\end{align*}
respectively, where $g(f(x_i^\prime))_j$ is the $j$-th element of $g(f(x_i^\prime))$, $\#\{x_i^\prime|P\}$ represents the count of $x_i^\prime$ satisfying the condition $P$.



\section{Results}
\label{results}
In this section, several comparative analyses evaluate CMCRL against classical contrastive learning methods (SimCLR \cite{ChenKN2020}, BYOL \cite{GrillSA2020}, and SimSiam \cite{ChenH2021}) and self-supervised algorithms based on clustering (CIKICS \cite{FangLL2021} and AutoEmbedder with K-means \cite{MonowarHK2022}).

In our experiments, we unify the size of all images to $256\times 256$. 
For the augmentation strategy $\mathcal{T}$, we set the probability of RHF as $0.5$, the parameter of Pad as $10$, and the filled values of RE as $[0.485, 0.456, 0.406]$. 
The embedding dimension $d$ is set as $512$. 
The number of instances and batch size are $4$ and $16$. 
The max neighbor distance and hyperparameters $k_1, k_2$ for the jaccard distance of DBSCAN is $0.4$, $30$, and $6$. 
The momentum $\alpha$ in Equation \eqref{momentum update} and the temperature $\tau$ in Equation \eqref{MLNCE} are set as $0.1$ and $0.05$. 
For the optimization process, the learning rate, weight decay, epoch count, and iteration count are $3.5\times 10^{-1}, 5\times 10^{-4}, 50$, and $100$. 
All experiments are implemented in the environment: Python 3.9, Pytorch 2.5.1, and CUDA 12.4 using a Windows machine with NVIDIA GeForce RTX 2070 and Intel(R) Core(TM) i7-10750H CPU @ 2.60GHz.

\begin{figure*}[!ht]
    \centering
    \includegraphics[width=\textwidth]{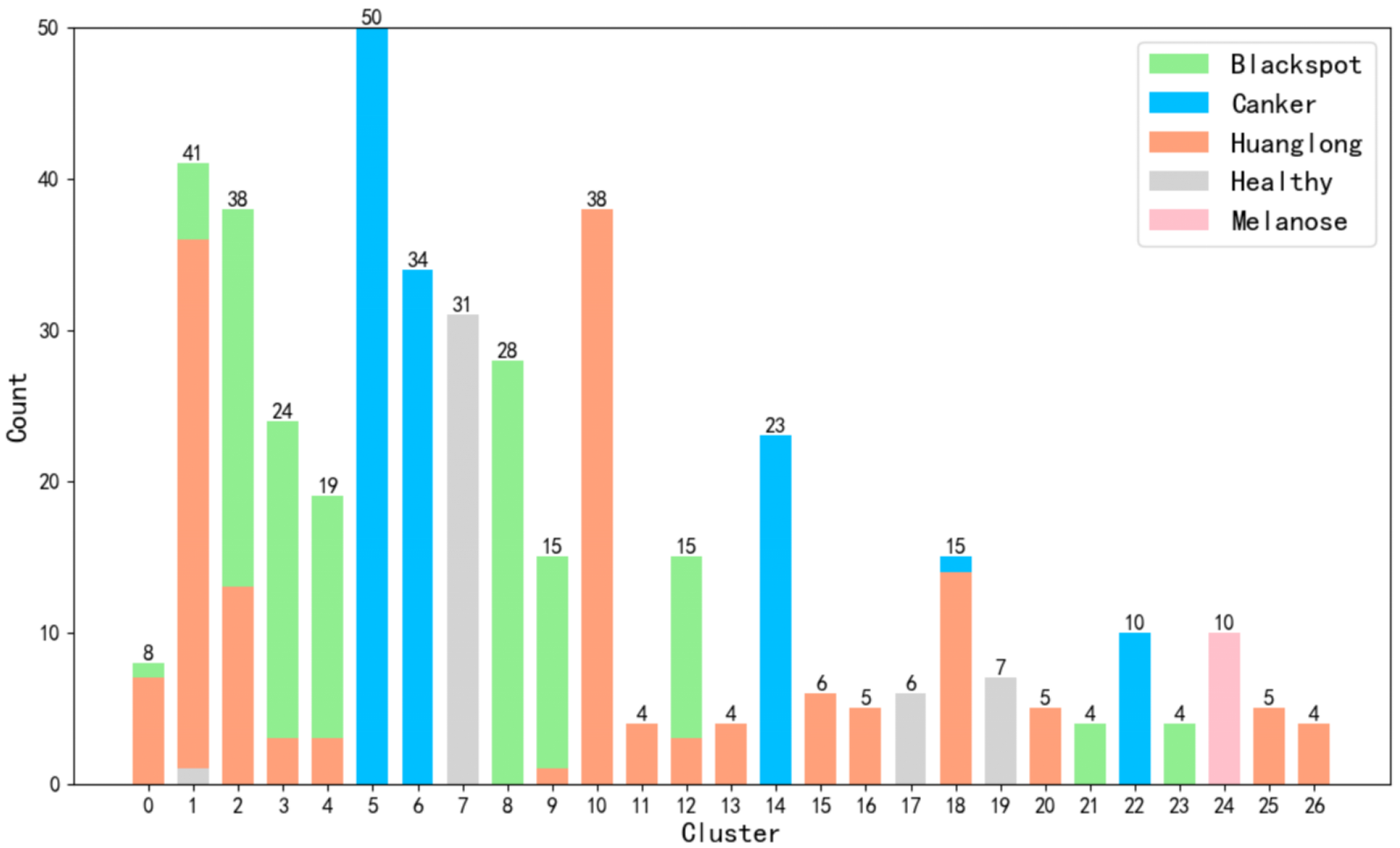}
    \caption{Cluster results of CMCRL.}
    \label{Cluster results}
\end{figure*}

\subsection{Comparison on clustering and accuracy}

\begin{table*}[!ht]
\caption{Comparison of downstream citrus disease classification accuracy (ACC ($\%$)) with the encoder ResNet50-IBN on CitrusDisease7.}
\label{Cluster Contrast vs NP vs MCT with ResNet50-IBN on CitrusDisease7}
    \centering
    \setlength{\tabcolsep}{1mm}
    \renewcommand\arraystretch{1.5}
    \begin{tabular}{@{\hspace{0.5cm}}c@{\hspace{0.5cm}}|@{\hspace{0.5cm}}c@{\hspace{0.5cm}}|@{\hspace{0.3cm}}c@{\hspace{0.3cm}}@{\hspace{0.3cm}}c@{\hspace{0.3cm}}@{\hspace{0.3cm}}c@{\hspace{0.3cm}}@{\hspace{0.3cm}}c@{\hspace{0.3cm}}@{\hspace{0.3cm}}c@{\hspace{0.3cm}}@{\hspace{0.3cm}}c@{\hspace{0.3cm}}@{\hspace{0.3cm}}c@{\hspace{0.3cm}}}
        \hline
        \multirow{2}{*}{Pre-training Method} & \multirow{2}{*}{Encoder} & \multicolumn{7}{c}{Epochs}\\
        & & 1 & 2 & 3 & 4 & 5 & 10 & 50\\
        \hline
        Cluster Contrast & \multirow{3}{*}{ResNet50-IBN} & 82.21 & 82.69 & 82.21 & 83.17 & 83.17 & 87.02 & 86.54\\
        Cluster Contrast + NP & & 82.46 & 82.69 & 86.54 & 86.54 & 85.10 & 87.98 & 87.50\\
        Cluster Contrast + MCT (CMCRL) & & \textbf{91.83} & \textbf{91.83} & \textbf{92.31} & \textbf{92.79} & \textbf{92.31} & \textbf{92.31} & \textbf{92.31}\\
        \hline
    \end{tabular}
\end{table*}

\begin{table*}[!ht]
\caption{Comparison of downstream citrus disease classification performance for several training paradigms on CitrusDisease7 and CDD. Specifically, “w.o. Pre-training” indicates Cluster Contrast initialized with random parameters. “ImageNet Pre-training” represents Cluster Contrast initialized with the ImageNet pre-training parameters. “Fully supervised Training” means that we use all 3956 labeled images to conduct supervised training rather than Cluster Contrast.}
\label{w.o. vs imagenet vs fully supervised}
    \centering
    \setlength{\tabcolsep}{1mm}
    \renewcommand\arraystretch{1.5}
    \begin{tabular}{@{\hspace{0.5cm}}c@{\hspace{0.5cm}}|@{\hspace{0.5cm}}c@{\hspace{0.5cm}}|@{\hspace{0.5cm}}c@{\hspace{0.5cm}}|@{\hspace{0.5cm}}c@{\hspace{0.5cm}}}
        \hline
        Training Paradigm & Encoder & Dataset & ACC ($\%$)\\
        \hline
        w.o. Pre-training & \multirow{4}{*}{ResNet50-IBN} & \multirow{4}{*}{CitrusDisease7} & 59.62\\
        ImageNet Pre-training & & & 83.65\\
        CMCRL Pre-training (This work) & & & 92.79\\
        Fully supervised Training & & & \textbf{93.27}\\
        \hline
        CMCRL Pre-training (This work) & \multirow{2}{*}{ResNet50-IBN} & \multirow{2}{*}{CDD} & \textbf{93.75}\\
        Fully supervised Training & & & 92.19\\
        \hline
    \end{tabular}
\end{table*}

\begin{table*}[!ht]
\caption{Comparison of other evaluation metrics of downstream citrus disease classification performance ($\%$) with the encoder ResNet50-IBN trained $10$ epochs on CitrusDisease7 and CDD.}
\label{Other metrices}
    \centering
    \setlength{\tabcolsep}{1mm}
    \renewcommand\arraystretch{1.5}
    \begin{tabular}{c|c|c|@{\hspace{0.5cm}}c@{\hspace{0.5cm}}@{\hspace{0.5cm}}c@{\hspace{0.5cm}}@{\hspace{0.5cm}}c@{\hspace{0.5cm}}}
        \hline
        \multirow{2}{*}{Pre-training Method} & \multirow{2}{*}{Encoder} & \multirow{2}{*}{Dataset} & \multicolumn{3}{c}{Metrics}\\
        & & & Recall & Precision & F1 Score \\
        \hline
        Cluster Contrast & \multirow{3}{*}{ResNet50-IBN} & \multirow{3}{*}{CitrusDisease7} & 78.58 & 77.37 & 76.85\\
        Cluster Contrast + NP & & & 81.55 & 80.50 & 79.80\\
        Cluster Contrast + MCT (CMCRL) & & & \textbf{85.67} & \textbf{84.39} & \textbf{84.14}\\
        \hline
        CIKICS \cite{FangLL2021} & ResNet50 & \multirow{4}{*}{CDD} & 42.20 & 52.10 & 46.60\\
        \cline{1-2}
        Cluster Contrast & \multirow{3}{*}{ResNet50-IBN} & & 88.62 & 89.82 & 87.12\\
        Cluster Contrast + NP & & & 89.86 & 91.25 & 89.30\\
        Cluster Contrast + MCT (CMCRL) & & & \textbf{94.75} & \textbf{94.43} & \textbf{93.50}\\
        \hline
    \end{tabular}
\end{table*}

\begin{table*}[!ht]
\caption{Comparison of downstream citrus disease classification performance (top-1 accuracy ($\%$)) with different encoders on the dataset CitrusDisease7.}
\label{Cluster Contrast vs NP vs MCT with other encoders on CitrusDisease7}
    \centering
    \setlength{\tabcolsep}{1mm}
    \renewcommand\arraystretch{1.5}
    \begin{tabular}{@{\hspace{0.5cm}}c@{\hspace{0.5cm}}|@{\hspace{0.5cm}}c@{\hspace{0.5cm}}|@{\hspace{0.5cm}}c@{\hspace{0.5cm}}@{\hspace{0.5cm}}c@{\hspace{0.5cm}}@{\hspace{0.5cm}}c@{\hspace{0.5cm}}@{\hspace{0.5cm}}c@{\hspace{0.5cm}}@{\hspace{0.5cm}}c@{\hspace{0.5cm}}@{\hspace{0.5cm}}c@{\hspace{0.5cm}}}
        \hline
        \multirow{2}{*}{Pre-training Method} & \multirow{2}{*}{Encoder} & \multicolumn{6}{c}{Epochs}\\
        & & 1 & 2 & 3 & 4 & 5 & 10\\
        \hline
        Cluster Contrast & \multirow{3}{*}{ResNet18} & 84.62 & 83.65 & 84.13 & 85.10 & 82.69 & 85.10\\
        Cluster Contrast + NP & & 86.54 & 86.54 & 87.98 & 88.46 & 87.02 & 88.46\\
        Cluster Contrast + MCT (CMCRL) & & \textbf{90.87} & \textbf{89.90} & \textbf{91.35} & \textbf{90.87} & \textbf{90.87} & \textbf{90.87}\\
        \hline
        Cluster Contrast & \multirow{3}{*}{ResNet34} & 75.00 & 76.44 & 79.81 & 78.85 & 77.40 & 81.73\\
        Cluster Contrast + NP & & 85.10 & 87.98 & 87.98 & 88.46 & 87.98 & 89.42\\
        Cluster Contrast + MCT (CMCRL) & & \textbf{89.42} & \textbf{88.94} & \textbf{91.83} & \textbf{91.35} & \textbf{90.87} & \textbf{92.79}\\
        \hline
        Cluster Contrast & \multirow{3}{*}{ResNet50} & 83.17 & 81.25 & 82.69 & 82.21 & 82.69 & 83.17\\
        Cluster Contrast + NP & & 83.54 & 84.13 & 86.54 & 86.54 & 82.21 & 85.58\\
        Cluster Contrast + MCT (CMCRL) & & \textbf{89.90} & \textbf{91.83} & \textbf{91.83} & \textbf{88.46} & \textbf{89.90} & \textbf{89.90}\\
        \hline
        Cluster Contrast & \multirow{3}{*}{ResNet101} & 75.96 & 79.81 & 78.85 & 76.92 & 78.85 & 81.73\\
        Cluster Contrast + NP & & 79.81 & 83.65 & 85.10 & 83.65 & 84.13 & 84.62\\
        Cluster Contrast + MCT (CMCRL) & & \textbf{89.90} & \textbf{88.46} & \textbf{88.46} & \textbf{88.46} & \textbf{87.02} & \textbf{89.90}\\
        \hline
        Cluster Contrast & \multirow{3}{*}{ResNet152} & 79.33 & 77.88 & 78.37 & 82.69 & 81.73 & 82.69\\
        Cluster Contrast + NP & & 83.17 & 85.10 & 81.73 & 81.25 & 84.62 & 83.65\\
        Cluster Contrast + MCT (CMCRL) & & \textbf{90.38} & \textbf{90.38} & \textbf{88.46} & \textbf{90.87} & \textbf{91.35} & \textbf{90.38}\\
        \hline
    \end{tabular}
\end{table*}

Table \ref{CMCRL vs CIKICS vs AutoEmbedder on CDD} compares the proposed method, CMCRL, with other classical self-supervised algorithms and clustering-based self-supervised algorithms in terms of clustering performance (CACC and ARI) and downstream classification accuracy, ACC. 
Compared with classical self-supervised algorithms (rows 1-3), CMCRL (last row) achieves superior ACC with a minimum performance gap 18.8\%. 
Meanwhile, CMCRL achieves 4.5\% and 29.7\% ACC enhancements over clustering-based self-supervised algorithms (rows 4,5), while yielding worse clustering scores (CACC=0.600 and ARI=0.228). 
This phenomenon stems from the contrastive mechanism of CMCRL. 
Specifically, Figure \ref{Cluster results} demonstrates that CMCRL generates 27 clusters, substantially exceeding the quantity of latent ground-truth labels, which directly correlates with CACC and ARI. 
The core mechanism in MLNCE loss (Equation \eqref{MLNCE}) enforces the instance-to-centroid contrast: maximizing the similarity between $z$ and its cluster centroid $\bar{z}^+$ while minimizing the similarity between $z$ and other cluster centroids $\bar{z}_j$. 
This design prioritizes intra-cluster semantic homogeneity over label-cluster alignment, fundamentally differentiating from pseudo-label dependency in conventional clustering-based approaches. 
As we can see, 26/27 (96.3\%) clusters in Figure \ref{Cluster results} present homogeneity (intra-cluster sample consistency $\geq$80\%), with only one cluster exhibiting cross-category mixture patterns. 
Therefore, CMCRL can establish enhanced feature representation through our proposed MCT paradigm. 

Table \ref{Cluster Contrast vs NP vs MCT with ResNet50-IBN on CitrusDisease7} presents the downstream classification accuracy of the classical Cluster Contrast, the variant with NP, and CMCRL pre-trained for 1-50 epochs on the dataset CitrusDisease7. 
These results reveal that MCT drives a 5.77\% ACC improvement (from 87.02\% to 92.79\%) over the original Cluster Contrast. 
Notably, MCT accelerates the pre-training process since CMCRL after single-epoch pre-training achieves comparable ACC to its optimal value, while Cluster Contrast requires ten epochs. 

As Table 1 of \cite{MonowarHK2022} demonstrated, AutoEmbedder with K-means underperforms supervised counterparts. 
Table \ref{w.o. vs imagenet vs fully supervised} studies the performance gap between CMCRL and the fully supervised paradigm on CitrusDisease7 and CDD. 
Specifically, for CitrusDisease7, we first provide the comparison between the CMCRL pre-training model (row 3) and the model with ImageNet pre-training parameters (row 2) and random initialization parameters (row 1). 
The corresponding ACC gaps are 9.14\% and 33.17\%, which confirms the effective knowledge transfer capability of CMCRL. 
Moreover, fully supervised training (row 4) has a marginal accuracy improvement (0.48\%) over CMCRL pre-training. 
While for CDD, the accuracy of our CMCRL pre-training model (row 5) even surpasses fully supervised training (last row) with an enhancement 1.56\%. 
Thus, it is reasonable to say that our methodology effectively bridges the performance disparity between self-supervised and supervised learning paradigms on fine-grained datasets. 

\subsection{Comparison on other metrics}
Except for ACC, we make comparison on other evaluation metrics, i.e., Recall, Precision, and F1 score. 
As far as we know, ACC measures the overall proportion of correct predictions relative to all samples, which is reliable for balanced class distributions. 
However, it becomes misleading for imbalanced datasets, as it prioritizes majority classes. 
To address this limitation, we employ F1 score, which combines Precision and Recall through harmonic averaging. 
Recall quantifies the detection completeness of true positive samples, while Precision evaluates the prediction validity among all positive samples. 
F1 score is designed to balance the inherent trade-off between Recall (excluding false positive samples) and Precision (excluding false negative samples), making it particularly suitable for class unbalanced datasets like CitrusDisease7 (unbalanced ratio 1: 29.4) and CDD (unbalanced ratio 1: 15.7). 

For CDD, Table \ref{Other metrices} shows that CMCRL obtains optimal behaviors (last row) on Recall (94.75\%), Precision (94.43\%), and F1 score (93.50\%). 
More than that, the F1 scores of Cluster Contrast (row 5) and the version with NP (row 6) achieve the improvements 40.52\% and 42.70\% over CIKICS (row 4), respectively. 
Although CIKICS employs the standard ResNet50 encoder rather than ResNet50-IBN, the ablation studies (Tables \ref{Cluster Contrast vs NP vs MCT with ResNet50-IBN on CitrusDisease7} and \ref{Cluster Contrast vs NP vs MCT with other encoders on CitrusDisease7} (rows 7-9)) demonstrate that the IBN modification yields so limited performance gains ($\leq 3\%$), insufficient to bring the aforementioned 40\% improvement. 
For CitrusDisease7, CMCRL (row 3 of Table \ref{Other metrices}) even provides about 7\% enhancement compared with Cluster Contrast (row 1) and its version with NP (row 2), separately achieving 85.67\%, 84.39\%, and 84.14\% for Recall, Precision, and F1 score. 
In summary, our proposed MCT paradigm can significantly improve the model performance on unbalanced dataset. 

\begin{table*}[!ht]
\caption{Comparison of downstream citrus disease classification performance (top-1 accuracy ($\%$)) with different augmentations on the dataset CitrusDisease7.}
\label{Cluster Contrast vs NP vs MCT with other augmentations on CitrusDisease7}
    \centering
    \setlength{\tabcolsep}{1mm}
    \renewcommand\arraystretch{1.5}
    \begin{tabular}{@{\hspace{0.5cm}}c@{\hspace{0.5cm}}|@{\hspace{0.5cm}}c@{\hspace{0.5cm}}|@{\hspace{0.5cm}}c@{\hspace{0.5cm}}@{\hspace{0.5cm}}c@{\hspace{0.5cm}}@{\hspace{0.5cm}}c@{\hspace{0.5cm}}@{\hspace{0.5cm}}c@{\hspace{0.5cm}}@{\hspace{0.5cm}}c@{\hspace{0.5cm}}@{\hspace{0.5cm}}c@{\hspace{0.5cm}}}
        \hline
        \multirow{2}{*}{Pre-training Method} & \multirow{2}{*}{Augmentation} & \multicolumn{6}{c}{Epochs}\\ & & 1 & 2 & 3 & 4 & 5 & 10\\
        \hline
        Cluster Contrast & \multirow{3}{*}{$\mathcal{T}/\{ \mathrm{RC}\}$} & 85.58 & 84.62 & 84.62 & 87.02 & 87.02 & 87.98\\
        Cluster Contrast + NP & & 85.10 & 86.06 & 83.17 & 89.90 & 87.98 & 87.50\\
        Cluster Contrast + MCT (CMCRL) & & \textbf{92.31} & \textbf{87.98} & \textbf{91.83} & \textbf{90.87} & \textbf{92.31} & \textbf{92.31}\\
        \hline
        Cluster Contrast & \multirow{3}{*}{$\mathcal{T}/\{ \mathrm{RE}\}$} & 83.65 & 84.13 & 82.21 & 84.13 & 84.62 & 87.98\\
        Cluster Contrast + NP & & 86.54 & 79.33 & 84.13 & 84.62 & 80.77 & 82.21\\
        Cluster Contrast + MCT (CMCRL) & & \textbf{91.83} & \textbf{92.31} & \textbf{90.38} & \textbf{92.31} & \textbf{89.90} & \textbf{93.75}\\
        \hline
        Cluster Contrast & \multirow{3}{*}{$\mathcal{T}/\{ \mathrm{Pad}\}$} & 80.77 & 83.65 & 82.69 & 82.21 & 85.58 & 86.06\\
        Cluster Contrast + NP & & 84.13 & 81.25 & 82.21 & 83.17 & 78.85 & 84.13\\
        Cluster Contrast + MCT (CMCRL) & & \textbf{92.31} & \textbf{89.90} & \textbf{91.35} & \textbf{90.38} & \textbf{90.38} & \textbf{91.83}\\
        \hline
        Cluster Contrast & \multirow{3}{*}{$\mathcal{T}/\{ \mathrm{RHF}\}$} & 85.58 & 81.73 & 79.33 & 86.54 & 84.13 & 83.17\\
        Cluster Contrast + NP & & 86.54 & 88.46 & 88.94 & 85.58 & 86.06 & 84.62\\
        Cluster Contrast + MCT (CMCRL) & & \textbf{91.35} & \textbf{90.87} & \textbf{92.31} & \textbf{94.23} & \textbf{89.42} & \textbf{90.87}\\
        \hline
        Cluster Contrast & \multirow{3}{*}{$\mathcal{T}/\{ \mathrm{RC,RE}\}$} & 83.65 & 84.62 & 84.62 & 85.58 & 83.65 & 83.65\\
        Cluster Contrast + NP & & 82.21 & 85.10 & 86.06 & 84.62 & 82.21 & 87.02\\
        Cluster Contrast + MCT (CMCRL) & & \textbf{92.31} & \textbf{92.79} & \textbf{91.35} & \textbf{93.27} & \textbf{88.94} & \textbf{89.42}\\
        \hline
        Cluster Contrast & \multirow{3}{*}{$\mathcal{T}/\{ \mathrm{Pad,RHF}\}$} & 83.65 & 86.06 & 83.65 & 82.21 & 81.73 & 79.33\\
        Cluster Contrast + NP & & 84.13 & 84.62 & 84.62 & 85.58 & 85.10 & 85.10\\
        Cluster Contrast + MCT (CMCRL) & & \textbf{92.79} & \textbf{87.98} & \textbf{91.83} & \textbf{93.27} & \textbf{89.42} & \textbf{88.94}\\
        \hline
    \end{tabular}
\end{table*}

\subsection{Comparison under different settings}
To validate the generalizability of our method, we conduct some ablation experiments with different experimental settings, such as encoder architectures and different augmentation strategies. 
Table \ref{Cluster Contrast vs NP vs MCT with other encoders on CitrusDisease7}, combined with Table \ref{Cluster Contrast vs NP vs MCT with ResNet50-IBN on CitrusDisease7}, presents the comparison among Cluster Contrast, the variant with NP, and CMCRL on multiple network architectures from ResNet18 to ResNet152. 
Results show two phenomenon: (1) the NP operation generally enhances 1\%-8\% ACC over the classical Cluster Contrast; (2) MCT paradigm further improves 2\%-7\% ACC. 
Furthermore, the acceleration effect of MCT is consistently demonstrated across all experimental settings. 
Finally, Table \ref{Cluster Contrast vs NP vs MCT with other augmentations on CitrusDisease7} shows that our method, CMCRL, achieves surprising performance improvements under various data augmentation strategies.

\section{Conclusions}
\label{conclusions}
This study develops a novel cluster-guided self-supervised learning framework, CMCRL, which classifies citrus diseases based on pre-training with a large amount of unannotated images and fine-tuning with small-scale annotated target images. 
Two key designs in this method, the contrastive learning with cluster centroids and the MCT paradigm, are proposed on the citrus disease dataset from two observations, respectively: (1) the symptom similarity across distinct citrus diseases; (2) hierarchical feature representations of neural networks. 
Unlike previous clustering-based self-supervised algorithms, CMCRL relaxes the requirement for clustering results, i.e., it only requires intra-cluster semantic homogeneity over
label-cluster alignment. 
Extensive experiments on the public dataset CDD and our proposed dataset CitrusDisease7 demonstrate that CMCRL achieves state-of-the-art classification performance compared with existing traditional self-supervised algorithms and clustering-based self-supervised algorithms, even comparable to fully supervised learning. 
Furthermore, CMCRL is beneficial to the model performance on unbalanced dataset.








\bibliographystyle{IEEEtran}
\bibliography{References}

\end{document}